\PassOptionsToPackage{usenames,dvipsnames}{xcolor}
\documentclass{article} 

\usepackage{iclr2023_conference,times}

\usepackage{amsmath,amsfonts,bm}

\def\eqref#1{equation~\ref{#1}}

\def\1{\bm{1}}

\DeclareMathAlphabet{\mathsfit}{\encodingdefault}{\sfdefault}{m}{sl}
\SetMathAlphabet{\mathsfit}{bold}{\encodingdefault}{\sfdefault}{bx}{n}

\usepackage{placeins}
\usepackage{hyperref}
\usepackage{url}
\usepackage{booktabs} 
\usepackage{graphicx}
\usepackage{cleveref}
\usepackage{wrapfig}
\usepackage{subcaption}
\definecolor{o1}{HTML}{fdae61}
\definecolor{b1}{HTML}{2b83ba}
\usepackage[dvipsnames]{xcolor}
\hypersetup{
	colorlinks,
	linkcolor={Thistle},
	citecolor={gray},
	urlcolor={TealBlue}
}
\newcommand{\MethodName}{\textit{LINR}}

\title{Revisiting Implicit Neural Representations in Low-Level Vision}

\author{Wentian Xu, Jianbo Jiao \\
School of Computer Science, University of Birmingham\\
\texttt{\{wxx155@alumni., j.jiao@\}bham.ac.uk} \\
}

\iclrfinalcopy 
\begin{document}

\maketitle

\begin{abstract}
Implicit Neural Representation (INR) has been emerging in computer vision in recent years. It has been shown to be effective in parameterising continuous signals such as dense 3D models from discrete image data, e.g. the neural radius field (NeRF). However, INR is under-explored in 2D image processing tasks. Considering the basic definition and the structure of INR, we are interested in its effectiveness in low-level vision problems such as image restoration.
In this work, we revisit INR and investigate its application in low-level image restoration tasks including image denoising, super-resolution, inpainting, and deblurring. Extensive experimental evaluations suggest the superior performance of INR in several low-level vision tasks with limited resources, outperforming its counterparts by over 2dB. 
Code and models are available at: \url{https://github.com/WenTXuL/LINR}.

\end{abstract}

\section{Introduction}
\label{sec:intro}

Implicit neural representation (INR) has been emerging in recent years. INR is a special type of continuous signal representation. It uses neural networks (typically multilayer perceptron, i.e. MLP) to parameterise complex signals such as 3D shapes as in NeRF~\citep{mildenhall2021nerf}. This representation has good storage efficiency and is widely used in 3D reconstruction because it can represent continuous output signals from discrete forms. Most of existing research on INR has been focused on 3D tasks, while it is under-explored in 2D low-level vision problems.
In this paper, we are interested in the question: \textit{How does INR perform in low-level vision?}

Low-level vision has always been a fundamental problem in computer vision. It mainly focuses on pixel-level problems such as image denoising, super-resolution, inpainting, and deblurring, to name a few. These image restoration tasks aim to obtain high-quality images from their corrupted versions. 
\begin{wrapfigure}{r}{0.5\textwidth}
\begin{center}
\includegraphics[width=0.9\linewidth]{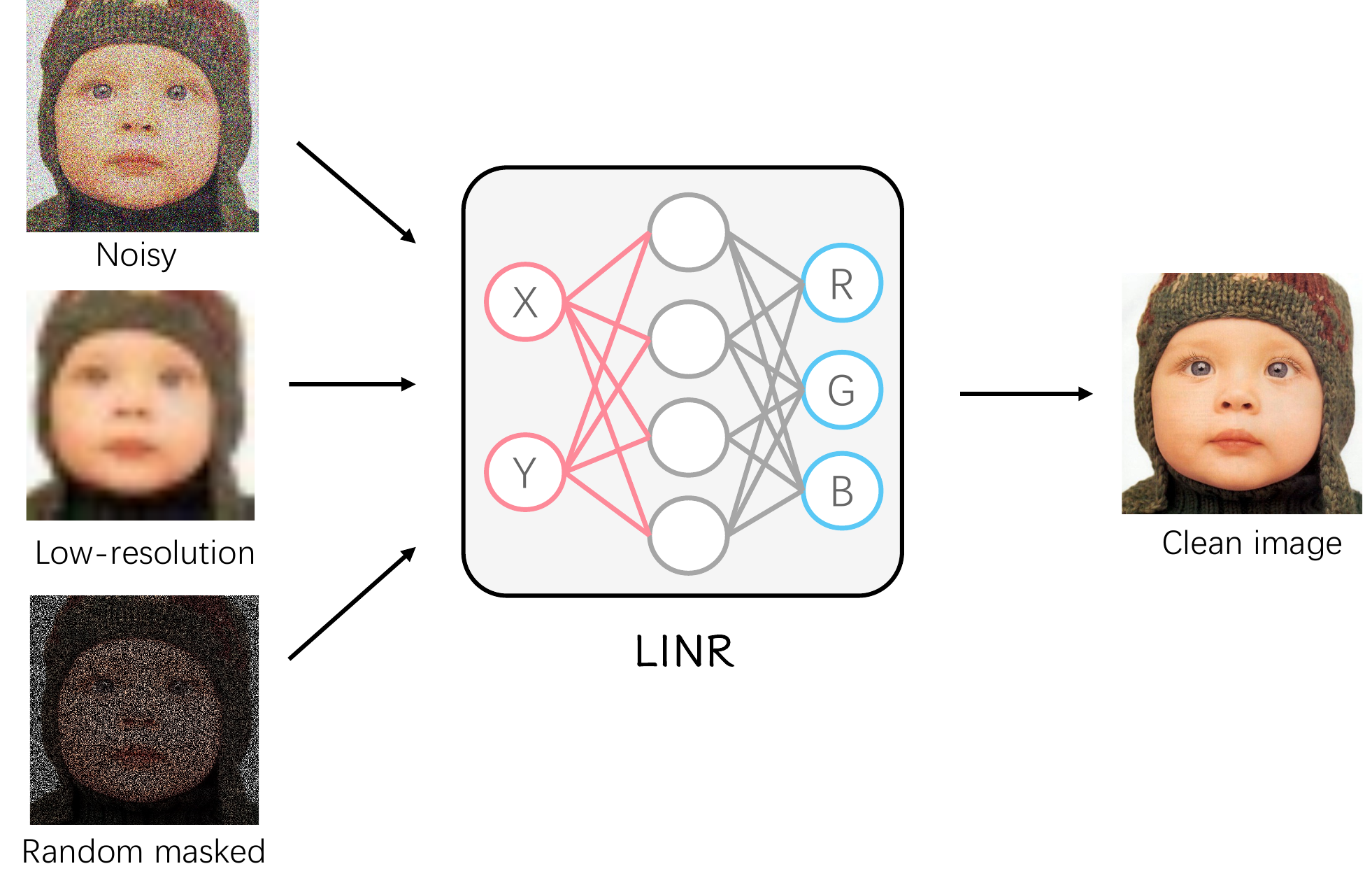}
\end{center}
\caption{The proposed \MethodName~aims to restore the original latent clean image (in terms of the RGB values of each pixel) from different types of corruptions, by only looking at the coordinates of the pixels within the input images.  
}
\vspace{-3mm}
\label{threeINR}
\end{wrapfigure}
With the development of computational resources, deep learning has been shown to be effective in addressing these tasks \citep{zhang2017beyond,zhang2018ffdnet,lai2017deep,yu2018generative},
while it usually requires large amounts of data with paired clean images for training the deep model. However, it is often difficult to obtain large-scale data with ground-truth clean images, and such methods tend to result in poor performance when the distribution of training data differs from that of the test data.

Therefore, image restoration based on a single image, i.e. obtaining a clean image from only one corrupted image without the ground-truth clean image, has become a promising solution in this field. 
Deep image prior (DIP)~\citep{ulyanov2018deep} investigates this solution by showing the potential of deep convolutional networks in addressing image restoration tasks without dataset-scale training.
 Some denoising works \citep{krull2019noise2void,lehtinen2018noise2noise} also get rid of the ground-truth clean image and  only need noisy images, but they still need a dataset to train the model. Self2Self \citep{quan2020self2self} follows these works by revising the network with dropout and mask, obtaining  a  high-quality clean image from a single noisy image. Zero-shot super-resolution (ZSSR) \citep{shocher2018zero} accomplishes super-resolution tasks with only one low-resolution image based on a fully convolutional network.

All these methods follow the conventional convolutional neural architecture, while the MLP (multi-layer perceptron) architecture that supports the INR is under-explored for low-level vision tasks. As a result, we are interested in its capability and performance in addressing low-level tasks, when given only one single image, similar to the aforementioned setting.

In our work, we propose to investigate the effectiveness of INR in low-level vision problems (\MethodName). 
We show the advantages of \MethodName~on four low-level image restoration tasks, including image denoising, super-resolution, inpainting, and deblurring. When given the same amount of limited resources, \MethodName~shows significantly better performance than other alternative approaches, validating the efficiency and effectiveness of the INR-based image restoration approach.

We further study the case where there are multiple corruptions by joint training the INR-based method. Surprisingly, \MethodName~shows much better performance than training with only one corruption, suggesting its effectiveness in extracting beneficial features from both corruptions during joint optimisation.
The main contributions of this work are summarised below:
\begin{itemize}
\item By revisiting INR in low-level vision tasks, we show the value and significant advantages of INR for 
image restoration tasks with limited resources.

\item 
The proposed \MethodName~is not restricted to one particular task, but generalises well to several different low-level vision tasks.

\item 
We show that INR-based approaches can significantly benefit from joint training with multiple corruptions. This is a bit counter-intuitive as conventional image restoration methods are usually confused given multiple types of corruption.

\end{itemize}
\section{Related Work}
\label{sec:related}

\paragraph{Implicit neural representation.}
Implicit neural representation (INR) has seen significant research interest in recent years. It focuses on parameterising a continuous differentiable signal with a neural network. This idea has been widely applied in 3D reconstruction tasks \citep{mildenhall2021nerf,mescheder2019occupancy,park2019deepsdf,saito2019pifu,sitzmann2019scene,oechsle2019texture,michalkiewicz2019implicit}. Whereas there is relatively fewer INR work on 2D image tasks, mainly focusing on image generation \citep{dupont2021generative,shaham2021spatially,skorokhodov2021adversarial,anokhin2021image}, and super-resolution \citep{chen2021learning,xu2021ultrasr}. For super-resolution tasks, We want to highlight that these methods require a large amount of data for training, whereas our \MethodName~ is concerned with cases where additional data is not required.
In addition, the traditional MLP with the ReLU activation function \citep{nair2010rectified} is difficult to parameterise image signals. \cite{sitzmann2020implicit} proposed SIREN to solve this problem.

\paragraph{Image restoration and Zero-shot image restoration.}
 As a low-level vision task, image restoration aims at obtaining a high-quality image from a corrupted image. Our work focuses on denoising, super-resolution, inpainting and deblurring. 
Most of the existing methods~\citep{lehtinen2018noise2noise,krull2019noise2void,batson2019noise2self,lai2017deep,shocher2018zero,pathak2016context,gao2017demand,mataev2019deepred} require large datasets for training, which in many cases are difficult to obtain. Researchers therefore propose methods that do not require extra data, also known as zero-shot methods. Self2self \citep{quan2020self2self} extends the idea of noise2void \citep{krull2019noise2void} by using masks and dropout to allow the model to learn enough information on a single noisy image. ZSSR \citep{shocher2018zero} can complete super-resolution tasks without any dataset. Most of these methods focus on a single task or a few tasks. However, deep image prior (DIP) \citep{ulyanov2018deep} shows that the convolutional generative model can be applied as a good image prior for most image restoration tasks without any additional training on data. DeepRED \citep{mataev2019deepred} has improved the DIP method based on RED \citep{romano2017little}. Our research also focuses on this zero-shot case. All of the previously mentioned methods are based on convolutional kernels. They also rely on some special structures like skip-connections. We show that, without these structures, \MethodName~ also performs well in these tasks.
\section{INR in low-level image restoration}
\label{sec:method}
Our work is based on implicit neural representation (INR). The aim of INR is to parameterise complex continuous signals by a neural network. In the case of a 2D colour image, INR is a mapping from the 2D coordinates system to the RGB value space: $f: \mathbb{R}^{2}\rightarrow \mathbb{R}^{3},f(x,y)=(r,g,b)$.

As opposed to conventional training strategies that rely on a large dataset, INR is able to accomplish image restoration tasks by parameterising image signals in only one image.
For instance, when we train our \MethodName~to represent noisy images, the noise will be inhibited in the early training iterations.

\begin{wrapfigure}{r}{0.5\textwidth}
\begin{center}
\vspace{-3mm}
\includegraphics[width=0.9\linewidth]{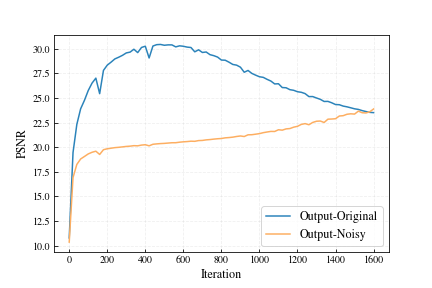}
\end{center}
\vspace{-3mm}
\caption{During training, the PSNR  between the model output and the ground-truth clean image (the \textcolor{b1}{blue} curve), and the PSNR between the output and the original noisy input (\textcolor{o1}{orange} curve). 
}
\label{denoiseshow}

\vspace{-2mm}
\end{wrapfigure}

As shown in \cref{denoiseshow}, 
the blue curve is the PSNR between the model output and the clean image. The orange curve is the PSNR between the output and the noisy input image. 
We can see that this training process suppresses the learning of noise. Although we only use the noisy image for training, the output approaches the clean image in the early iterations. Continued training will result in noisier results, but due to suppression, it is  hard for INR to learn the exact same input noisy image without enough capacity. 

In addition, we have found that the process of training INR to represent images can also complement the missing information in corrupted images, such as low-resolution images. We attribute all the above findings to the assumption that when training on the image, INR applies implicit restrictions on the output, including local continuity, global consistency, smoothness tendency in early iteration etc. These properties also exist in natural clean images, e.g. neighbouring pixels often change smoothly. As a result, the output that follows these restrictions is close to high-quality clean images. 
These restrictions are embedded in the structure and training process of INR.
For example, local continuity comes from continuous representation, part of global consistency comes from the sine activation function, and smoothness tendency comes from the training tendency of parameterised methods as these methods often tend to achieve smooth training at the beginning.

Based on this assumption, we propose the \MethodName~and
train on the corrupted image to achieve a high-quality clean target, which is achieved by the parameterising process. The implicit restriction hidden in this process will make sure the representation learned by \MethodName~ has the properties of a high-quality image while containing the information in the corrupted image. 
For different image restoration tasks, the only difference is the method of obtaining information from the corrupted image, i.e. the setting of the loss:
\begin{equation}
  \min\limits_{\theta}\left\| {f_{\theta}(c) - x_{0}} \right\|^{2}.
  \label{denoise}
\end{equation}

For the denoising task, as shown in \cref{denoise}, the \MethodName~ is simply trained on the corrupted image with the L2 loss, and then the output is the clean image in an early iteration. For the super-resolution task, we assume that the INR represents a high-resolution image and has the loss function:
\begin{equation}
  \min\limits_{\theta}\left\| {D(f_{\theta}(c)) - x_{0}} \right\|^{2},
  \label{SR}
\end{equation}
where $D(\cdot)$ is the downsampling method to get the low-resolution result from the \MethodName~output. Ideally, this low-resolution result should be similar to the original low-resolution image we actually have. In this case, the implicit restrictions of INR help with the reconstruction of details in the high-resolution image.

For the inpainting task, we assume that the \MethodName~represents a high-quality image and has the loss function:
\begin{equation}
  \min\limits_{\theta}\left\| {M(f_{\theta}(c)) -M(x_{0})} \right\|^{2},
  \label{inpainting}
\end{equation}
where $M(\cdot)$ is the masking function. As shown in \cref{inpainting}, we just need to ensure that the unobscured part of the original image is similar to the INR result, i.e. only the pixels without masks will be used for training. As a result, the \MethodName~training process will restore the obscured position.

The non-blind deblur task is similar to super-resolution and we have the loss function: 
\begin{equation}
  \min\limits_{\theta}\left\| {B(f_{\theta}(c)) - x_{0}} \right\|^{2},
  \label{blur}
\end{equation}
where $B(\cdot)$ is the blur kernel. The output of \MethodName~is blurred using the blur kernel, and then the blurred image is forced to be close to the original corrupted image as defined in the loss \cref{blur}.

Overall, \MethodName~is designed to represent and reconstruct high-quality clean images directly. Following that, its output is compared to the original corrupted image as a loss function to train the model.

\section{Experiment}
\label{sec:experiment}

In this section, we evaluate and investigate the effectiveness of the proposed \MethodName~over four low-level vision tasks, including image denoising, super-resolution, inpainting and deblurring. 
We also explore the performance of \MethodName~when facing more different corruptions.

The implementation of \MethodName~is based on SIREN \citep{sitzmann2020implicit}, which is a recent INR-based work replacing the activation function of the MLP with the sine function. This gives INR the ability to model information contained in higher-order derivatives, and the periodic feature of sine also provides SIREN with a degree of shift-invariance, which fits well with the properties of the image, as images often have repetitive information and feature.
The width of the MLP model
for all experiments is set to 256 and the depth to 6. The default learning rate is set to $10^{-4}$. The experiment is conducted on an NVIDIA GPU A100.

The parameters are initialised following SIREN. By default, the training iteration is the same (500 unless otherwise specified) for all the deep-learning methods and images in each task, for a more stable and fair comparison and to show the performance of image restoration with limited resources. 
PSNR and SSIM are used for the evaluation metrics.

\begin{table}[htbp]  
  \caption{Denoising performance (PSNR/SSIM) on Set9.}
\vspace{-3mm}
  \label{denoise25}
  \begin{center}
   \resizebox{0.87\linewidth}{!}{
    \begin{tabular}{lccc}
    \toprule
       Images   & DIP~\citep{ulyanov2018deep}   & S2S~\citep{quan2020self2self}   & \MethodName (Ours)\\
    \midrule
    Baboon & 19.68/0.502 & 20.29/0.581 & \textbf{23.67/0.794} \\
    F16   & 25.27/0.871 & 24.14/0.880 & \textbf{30.36/0.916} \\
    House & \textbf{28.34/0.887} & 24.31/0.851 & 23.10/0.580\\
    Lena  & 26.72/0.851 & 25.31/0.861 & \textbf{30.27/0.896} \\
    Peppers & 24.78/0.813 & 24.34/0.823 & \textbf{29.46/0.860} \\
    K01   & 22.19/0.624 & 22.40/0.711 & \textbf{25.99/0.818} \\
    K02   & 27.83/0.834 & 26.21/0.829 & \textbf{29.66/0.866} \\
    K03   & 27.63/0.869 & 24.77/0.860 & \textbf{30.40/0.903} \\
    K12   & 27.47/0.858 & 24.93/0.864 & \textbf{30.34/0.892} \\
    \midrule
    Average & 25.55/0.790 & 24.08/0.807 & \textbf{28.14/0.836}\\
    \bottomrule
    \end{tabular}
    }
    \end{center}
\end{table}

\subsection{Denoising}
\begin{figure}[ht!]
\centering
	\begin{subfigure}{.9\textwidth}		      
	\centering
	\includegraphics[width=\textwidth]{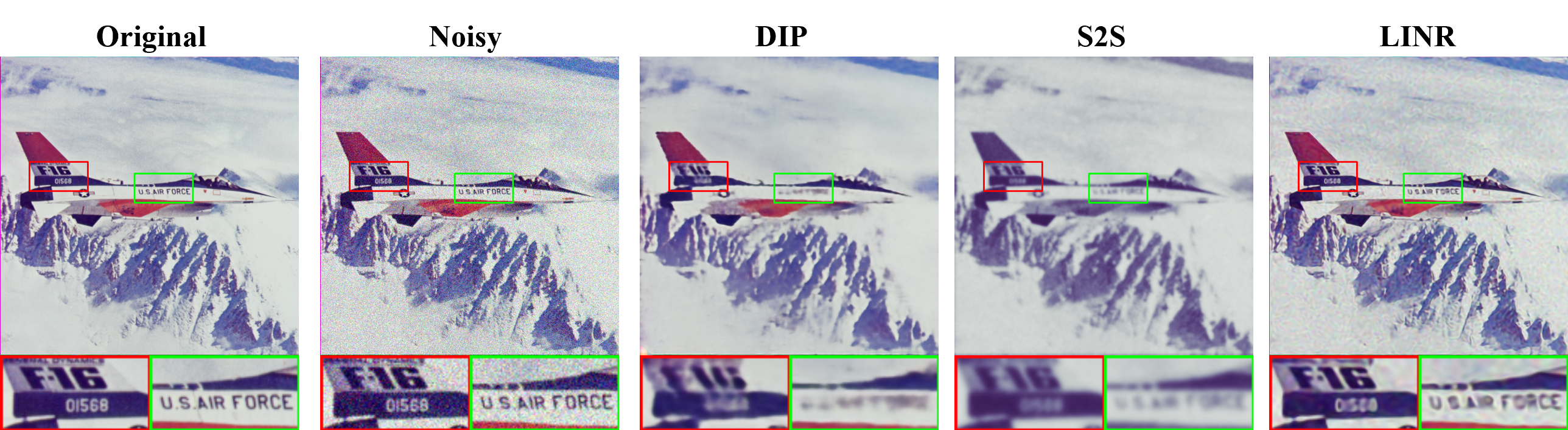}
			\caption{Denoising}
                \label{qa(a)}
		\end{subfigure}
		
		\begin{subfigure}{.9\textwidth}
			\centering
			\includegraphics[width=\textwidth]{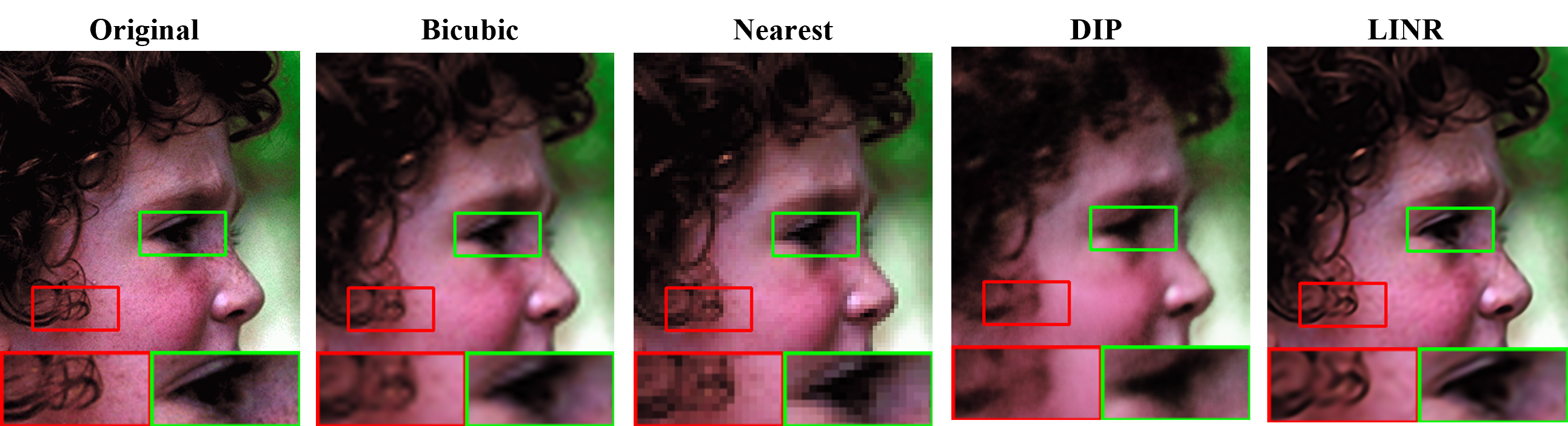}
			\caption{Super-resolution}
                \label{qa(b)}
		\end{subfigure}
            \begin{subfigure}{.9\textwidth}
			\centering
			\includegraphics[width=\textwidth]{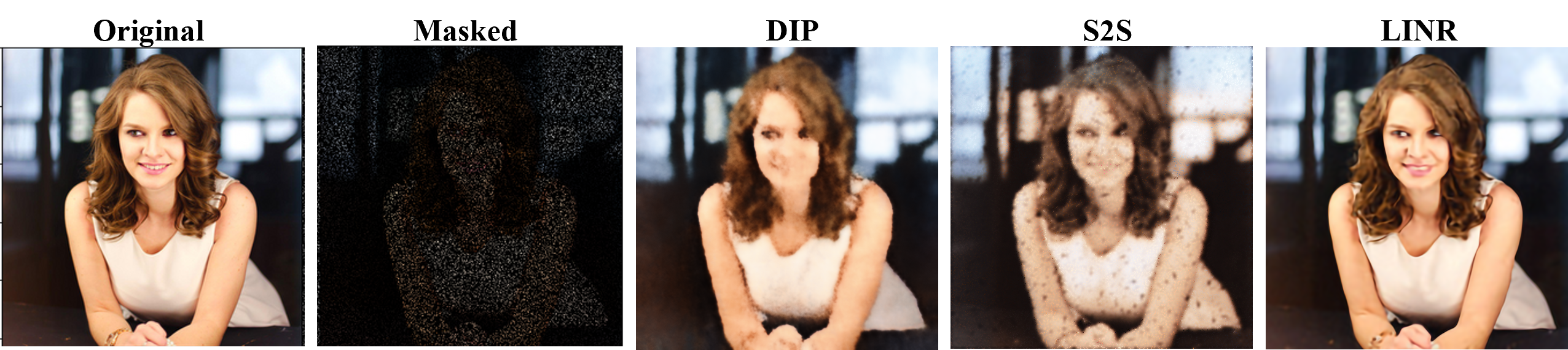}
			\caption{Inpainting}
                \label{qa(c)}
		\end{subfigure}
            \begin{subfigure}{.9\textwidth}
			\centering
			\includegraphics[width=\textwidth]{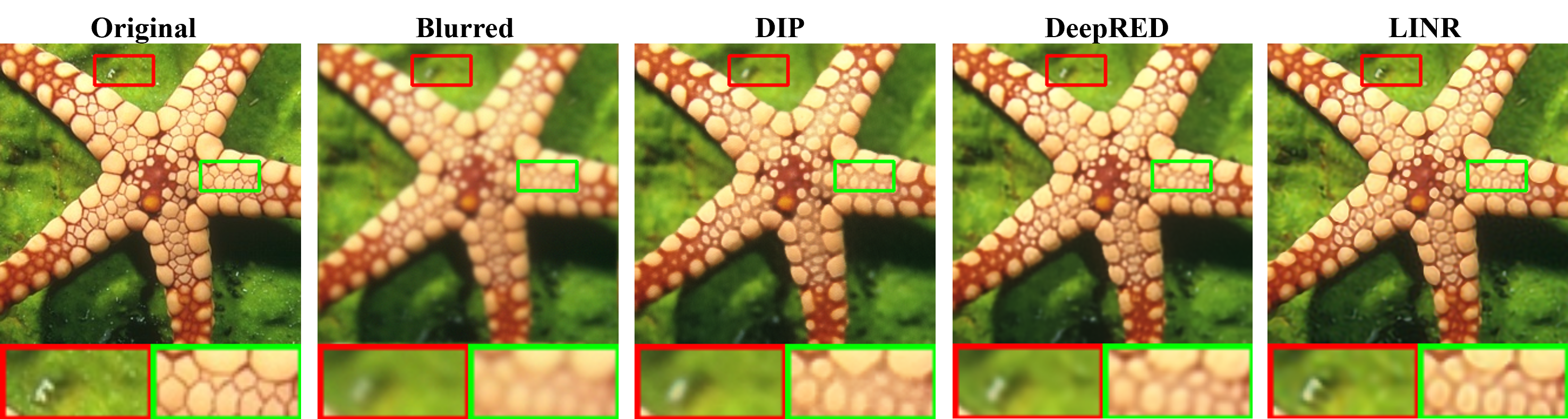}
			\caption{Deblurring}
                \label{qa(d)}
		\end{subfigure}

\caption{Qualitative results of different tasks with comparison to state-of-the-arts.} 
\label{denoisefigure}
\end{figure}

First, we show the performance of \MethodName~on image denoising. We use the Set9 dataset 
as in \citep{ulyanov2018deep} and add Gaussian noise with $\sigma = 25$. The result is shown in \cref{denoise25} and \cref{qa(a)}. It can be seen that the average results of \MethodName~are significantly better than the other methods for both SSIM and PSNR. 
As mentioned before, \MethodName~needs to stop early in the training period for denoising, and the best stopping position is different for different images. This is especially the case if the size of the image is different. However, the same stopping point is chosen for all images and methods for a fair comparison. So \MethodName's results are poor on smaller-size images like the house (256*256). INR is so efficient that it learns to represent the clean version very quickly on these small images and then moves on to learn more noise information. 
The training time for LINR on a 256*256 image is 8.3s, and 14.3s for DIP as a comparison. For higher-resolution images, LINR will take more time as it is trained on all pixels with MLP.

\subsection{Super-resolution}
We then conduct experiments for image super-resolution on Set5 \citep{bevilacqua2012low} with a sampling factor of 4. The downsampling method used in the loss function is lanczos2 \citep{turkowski1990filters}. The results are shown in \cref{super-resolution set5} and \cref{qa(b)}, from which we can see that the \MethodName~performs significantly better than the other methods both quantitatively and qualitatively.

\subsection{Inpainting}

For the inpainting experiments, the mask is generated randomly.
We show the results on sparsity 0.1. This means only ten percent of pixels remain after the masking. Again, \MethodName~performs significantly better than the other methods, as shown in \cref{inpaintingtest} and \cref{qa(c)}.

\begin{minipage}[t]{0.5\textwidth}
\makeatletter\def\@captype{table}
\caption{Super-resolution results on Set5 with factor = 4.}
\resizebox{\linewidth}{!}{

\begin{tabular}{lcccc}
   \toprule
       Images    & Bicubic & Nearest & DIP~\citep{ulyanov2018deep}   & \MethodName \\
    \midrule
    Baby  & 30.73/0.916 & 27.76/0.875 & 26.40/0.827 & \textbf{31.20/0.921} \\
    Bird & 28.44/0.917 & 25.22/0.851 & 25.57/0.797 & \textbf{30.48/0.944} \\
    Butterfly & 21.18/0.797 & 18.85/0.725 & 22.68/0.857 & \textbf{24.29/0.898} \\
    Head  & 29.08/0.841 & 27.79/0.811 & 26.58/0.768 & \textbf{29.58/0.854} \\
    Woman & 25.39/0.885 & 22.90/0.827 & 24.28/0.837 & \textbf{27.44/0.924} \\
    \midrule
    Average & 26.96/0.871 & 24.50/0.818 & 25.10/0.817 & \textbf{28.60/0.908} \\
    \bottomrule
\end{tabular}
}

\label{super-resolution set5}
\end{minipage}
\begin{minipage}[t]{0.5\textwidth}
\makeatletter\def\@captype{table}
\caption{Inpainting results on images with sparsity of 0.1 Bernoulli random masks.}
\resizebox{\linewidth}{!}{
\begin{tabular}{lccc}
     \toprule
        Images  & DIP~\citep{ulyanov2018deep}   & S2S~\citep{quan2020self2self}   & \MethodName \\
    \midrule
    Kate  & 25.87/0.893 & 22.60/0.841 & \textbf{31.78/0.956} \\
    Baby  & 25.99/0.826 & 24.36/0.802 & \textbf{29.27/0.903} \\
    Bird & 24.88/0.800 & 21.78/0.736 & \textbf{28.21/0.922} \\
    Butterfly & 20.59/0.827 & 18.04/0.706 & \textbf{21.05/0.853} \\
    Head  & \textbf{26.53}/0.770 & 22.94/0.743 & 26.39/\textbf{0.775} \\
    Woman & 23.53/0.840 & 20.35/0.747 & \textbf{24.41/0.892} \\
    \midrule
    Average & 24.57/0.826 & 21.68/0.763 & \textbf{26.85/0.884} \\
    \bottomrule
\end{tabular}
}
\label{inpaintingtest}

\end{minipage}

\subsection{Deblurring}
For the deblurring task, we show the results for the Gaussian blur in \cref{deblurtest} and \cref{qa(d)}. A Gaussian kernel of width 25 and sigma 1.6 was chosen as the blur kernel. We can see that \MethodName~performs better than the other methods. 
In this task, the iteration was set to 4000, as a lower number of iterations results in unreasonable results.

\begin{table}[htbp]
  
  \caption{Deblurring results (PSNR) on Gaussian blurred images ($\sigma=1.6$, width = 25).}
  \vspace{-3mm}
  \begin{center}
   \resizebox{0.81\linewidth}{!}{
    \begin{tabular}{lccc}
    \toprule
     Images     & DIP~\citep{ulyanov2018deep} & DeepRED~\citep{mataev2019deepred} & \MethodName \\
    \midrule
    Leaves & 28.15  & 29.85  & \textbf{30.10}  \\
    Parrot & 30.03  & 31.30  & \textbf{32.84}  \\
    Starfish & 28.99  & 30.08  & \textbf{31.78}  \\
    Butterfly & 28.96  & \textbf{30.15}  & 30.07  \\
    \midrule
    Average & 29.03  & 30.35  & \textbf{31.20}  \\
    \bottomrule
    \end{tabular}
    }
  \label{deblurtest}
  \end{center}
  \vspace{-3mm}
\end{table}

\subsection{Joint training}
Here we explore the performance of \MethodName~under multiple types of different corruption. The models are jointly trained with two corrupted images, which are derived from the same original image but with different corruption types. We found that \MethodName~performs very well on joint training tasks. It can combine information from different corrupted images while not being misled by the differences between the corruptions.
We tested joint training on super-resolution, denoising, and inpainting. The average PSNR performance of joint training with denoising and 
super-resolution tasks with limited resources is 30.98, outperforming DIP by a large margin (25.06). More importantly, it is significantly higher than training on a single corruption (Denoising: 27.63,  SR: 28.60 ). Due to the space limit, detailed comparison results of other joint-training tasks are presented in the appendix~\ref{appdix}.

The core idea of image restoration based on INR is to guarantee the high quality of the represented image through implicit restrictions rather than learning the direct mapping between corrupted images and original images. This core process is independent of the corruption type. \MethodName~is therefore not limited to one image restoration task and will not be misled by the differences between corruptions. More corrupted images will provide richer information to help \MethodName~express high-quality images. Our finding shows the feasibility of learning under multiple different corruptions with INR.

\section{Conclusion}
\label{sec:conclusion}
In this paper, we revisited INR and explored its applications in several image restoration tasks. Under the same setting, it has been shown that the proposed \MethodName~significantly outperforms other alternative solutions across all the evaluated tasks, suggesting the effectiveness of such simple representations in low-level tasks. Furthermore, we showcased the strong performance of \MethodName~when given more realistic multiple corruptions, outperforming its single-corruption counterparts as well as other competitive solutions. We believe the findings derived from this study will potentially attract research interest in this new direction.

Similar to DIP \citep{ulyanov2018deep}, one possible limitation of \MethodName~ is that the denoising process is easy to overfit. Therefore, the study of the stop criterion (e.g. \citep{jo2021rethinking}) could be a future direction worth investigating.

\bibliography{iclr2023_conference}
\bibliographystyle{iclr2023_conference}

\newpage
\appendix
\section{Appendix}\label{appdix}
\hyperref[sec1]{\textbf{1. Table of Joint-training}}~\\~\\
\hyperref[sec2]{\textbf{2. An ablation study with different activation functions}}~\\~\\
\hyperref[sec3]{\textbf{3. Super-resolution results with different factors}}~\\~\\
\hyperref[sec4]{\textbf{4. Qualitative results on joint training}}~\\~\\
\hyperref[sec5]{\textbf{5. Qualitative results on inpainting}}~\\~\\
\hyperref[sec6]{\textbf{6. Qualitative results on super-resolution}}~\\~\\
\hyperref[sec7]{\textbf{7. Qualitative results on deblurring}}~\\~\\
\hyperref[sec8]{\textbf{8. Additional qualitative results and real noisy image denoising}}~\\~\\

\subsection{Table of Joint-training}
\label{sec1}
\FloatBarrier
Here, We show the results of the combination of super-resolution and denoising in \cref{denoisesr}, denoising and 
inpainting in \cref{denoisein}, and super-resolution and inpainting in \cref{srin}. In these  settings, the performance of \MethodName~is significantly better than its counterpart.

Furthermore, in \cref{m22}, \cref{m21} and \cref{m23}, we compare joint training and training on only one corruption. 
It can be seen that joint training achieves significantly better performance than training on a single corruption across all three tasks, suggesting the effectiveness of the \MethodName.

For images containing noise in all joint-training experiments, the training weight is set to 0.1. This aims to reduce the misleading effect of this type of corruption. For all other images, the weight is 1.

\begin{table}[htbp]
  
  \caption{The restoring results of joint training on noisy and low-resolution images.}
  \begin{center}
    \begin{tabular}{lcc}
    \toprule
       Images   & DIP~\citep{ulyanov2018deep}   & \MethodName \\
          \midrule
    Baby  & 25.96/0.823 & \textbf{31.80/0.927} \\
    Bird & 25.74/0.804 & \textbf{33.28/0.962} \\
    Butterfly & 23.32/0.873 & \textbf{28.89/0.950} \\
    Head  & 26.63/0.772 & \textbf{30.07/0.864} \\
    Woman & 23.63/0.833 & \textbf{30.86/0.955} \\
    \midrule
    Average & 25.06/0.821 & \textbf{30.98/0.932} \\
    \bottomrule
    \end{tabular}
  \label{denoisesr}
  \end{center}
\end{table}

\begin{table}[htbp]
  
  \caption{The restoring results of joint training on noisy and random masked (inpainting) images.}
  \begin{center}    
  \begin{tabular}{lcc}
    \toprule
      Images    & DIP~\citep{ulyanov2018deep}   & \MethodName \\
          \midrule
    Kate  & 25.93/0.893 & \textbf{33.41/0.964} \\    
    F16   & 25.52/0.871 & \textbf{31.08/0.942} \\
    Peppers & 25.01/0.825 & \textbf{30.62/0.891} \\  
    House & 28.29/\textbf{0.890} & \textbf{29.00}/0.843 \\  
    Baboon & 19.47/0.502 & \textbf{24.05/0.813} \\
    Lena  & 26.78/0.852 & \textbf{31.49/0.919}\\
    k01   & 22.03/0.624 & \textbf{25.71/0.807} \\
    k02   & 28.10/0.838 & \textbf{29.66/0.871} \\
    k03   & 27.48/0.869 & \textbf{30.98/0.914} \\
    k12   & 27.55/0.857 & \textbf{30.55/0.896} \\
    
    \midrule
    Average & 25.62/0.802 & \textbf{29.65/0.886} \\
    \bottomrule
    \end{tabular}
  \label{denoisein}
  \end{center}
\end{table}

\begin{table}
 
  \caption{The restoring results of joint training on low-resolution and random masked (inpainting) images.}
  \begin{center}
    \begin{tabular}{lcc}
    \toprule
       Images   & DIP~\citep{ulyanov2018deep}   & \MethodName \\
          \midrule
    Baby  & 25.83/0.821 & \textbf{31.57/0.924} \\
    Bird & 25.02/0.797 & \textbf{32.58/0.962} \\
    Butterfly & 23.14/0.869 & \textbf{26.36/0.934} \\
    Head  & 26.30/0.766 & \textbf{29.71/0.861} \\
    Woman & 24.33/0.847 & \textbf{29.47/0.948} \\
    \midrule
    Average & 24.92/0.820 & \textbf{29.94/0.926} \\
    \bottomrule
    \end{tabular}
  \label{srin}
  \end{center}
\end{table}

\begin{table}
  \caption{The \MethodName~joint results where both noisy and low-resolution images are used as input and the \MethodName~restoring results where only one of them is used.}
    \begin{center}
    \begin{tabular}{lccc}
    \toprule
       Images   & Denoising & SR    & Together \\
          \midrule
   Baby  & 30.58/0.915 & 31.20/0.921 & \textbf{31.80/0.927} \\
    Parrot & 27.14/0.796 & 30.48/0.944 & \textbf{33.28/0.962} \\
    Butterfly & 24.82/0.821 & 24.29/0.898 & \textbf{28.89/0.950} \\
    Head  & 27.36/0.765 & 29.58/0.854 & \textbf{30.07/0.864} \\
    Women & 28.24/0.868 & 27.44/0.924 & \textbf{30.86/0.955} \\
    \midrule
    Average & 27.63/0.833 & 28.60/0.908 & \textbf{30.98/0.932} \\
    \bottomrule
    \end{tabular}
  \label{m22}
  \end{center}
\end{table}

\begin{table}
  \begin{center}
  \caption{The \MethodName~joint training results where both noisy and random masked (inpainting) images are used as input and the \MethodName~restoring results where only one of them is used.}
    \begin{tabular}{lccc}
    \toprule
       Images   & Denoising & Inpainting & Together \\
    \midrule
    Kate  & 32.00/0.951 & 31.78/0.956 & \textbf{33.41/0.964} \\
    F16   & 30.36/0.916 & 26.00/0.901 & \textbf{31.08/0.942} \\
    Pepper & 29.46/0.860 & 27.30/0.879 & \textbf{30.62/0.891} \\
    House & 23.10/0.580 & 26.29/0.862 & \textbf{29.00/0.843} \\
    Baboon & 23.67/0.794 & 17.43/0.508 & \textbf{24.05/0.813} \\
    Lena  & 30.27/0.896 & 27.84/0.887 & \textbf{31.49/0.919} \\
    k01   & \textbf{25.99/0.818} & 19.91/0.607 & 25.71/0.807 \\
    k02   & 29.66/0.866 & 28.31/0.857 & \textbf{29.66/0.871} \\
    k03   & 30.40/0.903 & 29.02/0.902 & \textbf{30.98/0.914} \\
    k12   &30.34/0.892 & 28.34/0.880 & \textbf{30.55/0.896} \\
    \midrule    
    Average & 28.53/0.848 & 26.22/0.824 & \textbf{29.65/0.886} \\
    \bottomrule
    \end{tabular}
  \label{m21}
  \end{center}
\end{table}

\begin{table}
   \caption{The \MethodName~joint results where both low-resolution and random masked (inpainting) images are used as input and the \MethodName~restoring results where only one of them is used.}
   \begin{center}
    \begin{tabular}{lccc}
    \toprule
       Images   & Inpainting & SR    & Together \\
          \midrule
    Baby  & 29.27/0.903 & 31.20/0.921 & \textbf{31.57/0.924} \\
    Parrot & 28.21/0.922 & 30.48/0.944 & \textbf{32.58/0.962} \\
    Butterfly & 21.05/0.853 & 24.29/0.898 & \textbf{26.36/0.934} \\
    Head  & 26.39/0.775 & 29.58/0.854 & \textbf{29.71/0.861} \\
    Women & 24.41/0.892 & 27.44/0.924 & \textbf{29.47/0.948} \\
    \midrule
    Average & 25.87/0.869 & 28.60/0.908 & \textbf{29.94/0.926} \\
    \bottomrule
    \end{tabular}
  \label{m23}
  \end{center}
\end{table}
\FloatBarrier
\subsection{An ablation study with different activation functions}
\FloatBarrier
\label{sec2}

Our work is based on SIREN \citep{sitzmann2020implicit}, which uses sine as the activation function. We also replace it with other activation functions for super-resolution tasks as an ablation study here. We have tried Tanh, Sigmoid, ReLU, etc. We have also tried using positional encoding with ReLU, like what is used in NeRF \citep{mildenhall2021nerf}. The result is shown in \cref{srablation}. With other activation functions, INR performed poorly on SR with the same number of training iterations(500). SIREN can represent high-quality images in a very short time, whereas the other activation functions can not. If given more training iterations, the performance of ReLU-PE can increase significantly (with an average PSNR of 26.23 on Set5 after 2000 iterations), but is still much worse than the result of SIREN (because ReLU cannot model higher-order derivatives and also lack the ability to extract enough global information \citep{sitzmann2020implicit}.) However, it is significantly better than the result of Nearest (24.50), which shows that with other activation functions, INR also has the capacity to restore high-quality images.
\begin{table}[htbp]
  \centering
  \caption{Super-resolution results of INR on Set5 with different activation functions.}
  \resizebox{\linewidth}{!}{
    \begin{tabular}{lcccccc}
    \toprule
          & Sigmoid   & Tanh   & ReLU  & SeLU & ReLU-PE & Sine (SIREN/\MethodName) \\
    \midrule
    
    Set5 & 14.48/0.520 & 15.29/0.520 & 19.94/0.643 & 18.18/0.590 & 24.29/0.742 & 28.60/0.908\\
    \bottomrule
    \end{tabular}
    }
  \label{srablation}
\end{table}
\FloatBarrier

\subsection{Super-resolution results with different factors}
\FloatBarrier
\label{sec3}
\begin{table}[htbp]
  \centering
  \caption{Super-resolution results (PSNR) on Set5 with factor = 8.}
    \begin{tabular}{lcccc}
    \toprule
          & Bicubic & Nearest & DIP~\citep{ulyanov2018deep} & \MethodName \\
          \midrule
    Baby  & 26.14  & 24.03  & 24.53  & \textbf{27.32 } \\
    Bird  & 23.30  & 21.48  & 22.30  & \textbf{24.07 } \\
    Butterfly & 16.85  & 15.67  & 18.16  & \textbf{18.66 } \\
    Head  & 26.99  & 25.50  & 25.18  & \textbf{27.48 } \\
    Woman & 21.25  & 19.60  & 21.09  & \textbf{22.43 } \\
    \midrule
    Average & 22.91  & 21.26  & 22.25  & \textbf{23.99 } \\
    \bottomrule
    \end{tabular}

  \label{sr8}
\end{table}

\begin{table}[htbp]
  \centering
  \caption{Super-resolution results (PSNR) on Set5 with factor = 2.}
    \begin{tabular}{lcccc}
    \toprule
          & Bicubic & Nearest & DIP~\citep{ulyanov2018deep} & \MethodName \\
          \midrule
    Baby  & \textbf{36.02 } & 32.64  & 26.63  & 32.87  \\
    Bird  & 35.56  & 30.53  & 26.15  & \textbf{36.86 } \\
    Butterfly & 26.59  & 23.43  & 24.63  & \textbf{28.82 } \\
    Head  & \textbf{31.67 } & 30.54  & 27.18  & 31.08  \\
    Woman & 31.34  & 27.75  & 25.56  & \textbf{32.75 } \\
    \midrule
    Average & 32.24  & 28.98  & 26.03  & \textbf{32.48 } \\
    \bottomrule
    \end{tabular}
  \label{sr2}
\end{table}

\begin{figure}[h]
\begin{center}
\includegraphics[width=1.0\linewidth]{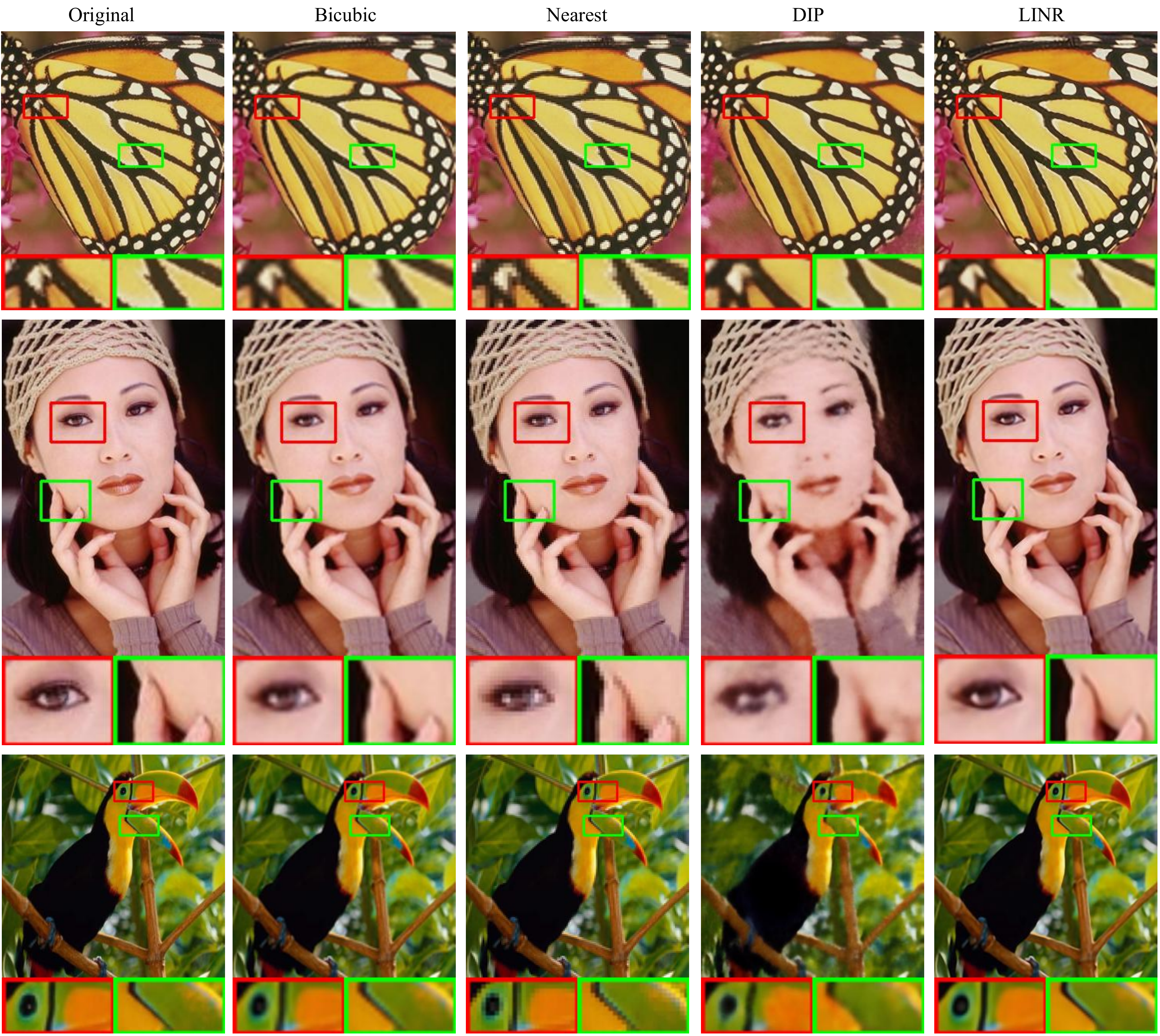}
\end{center}
\caption{Super-resolution results on Set5 with factor = 2.  }
\label{SRfig2}
\end{figure}

\begin{figure}[h]
\begin{center}
\includegraphics[width=1.0\linewidth]{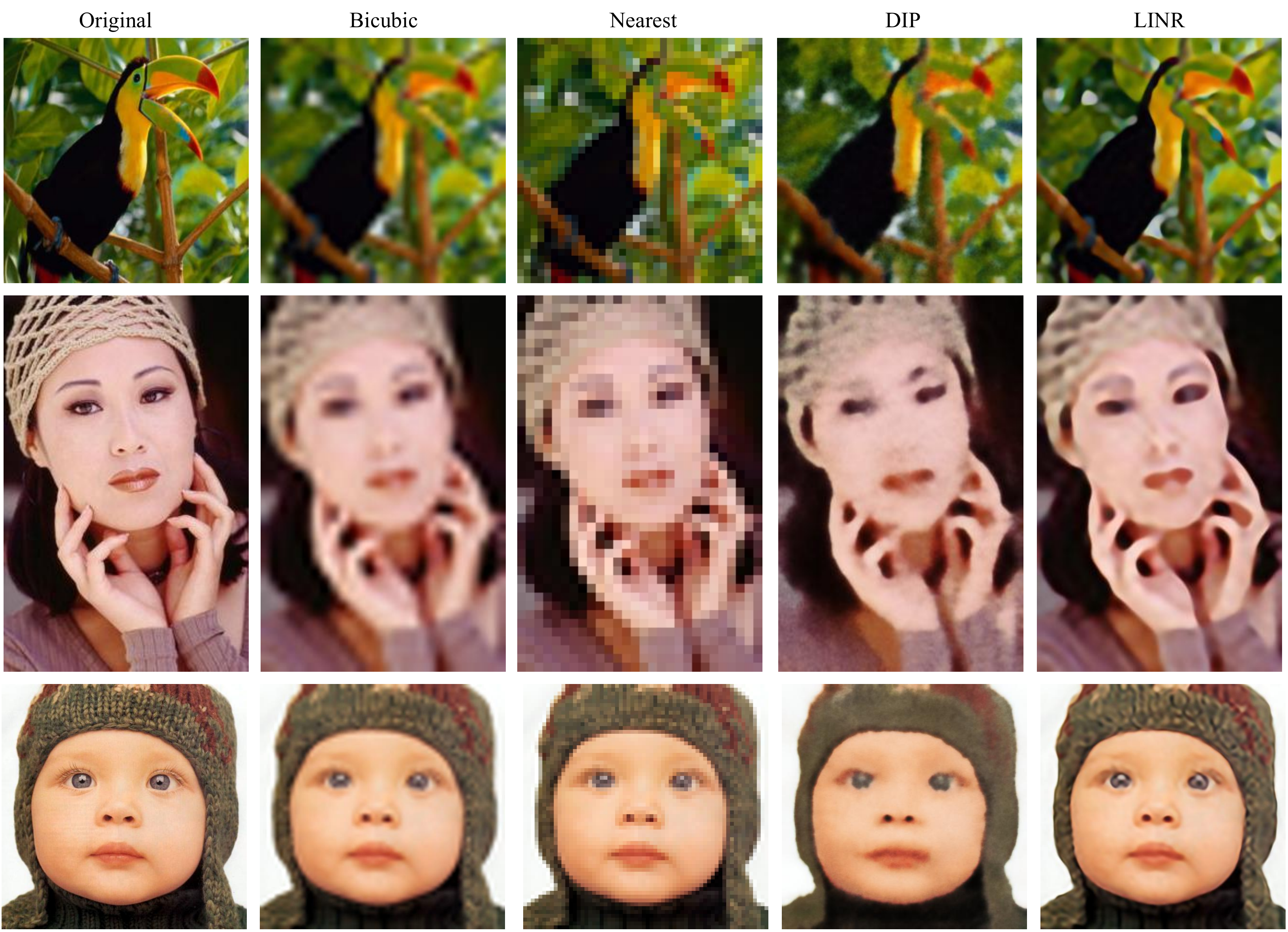}
\end{center}
\caption{Super-resolution results on Set5 with factor = 8.  }
\label{SRfig8}
\end{figure}
\newpage
\FloatBarrier

\newpage
\subsection{Qualitative results on joint training }
\FloatBarrier
\label{sec4}
~\\~\\~\\~\\~\\
\begin{figure}[h]
\begin{center}
\includegraphics[width=1.0\linewidth]{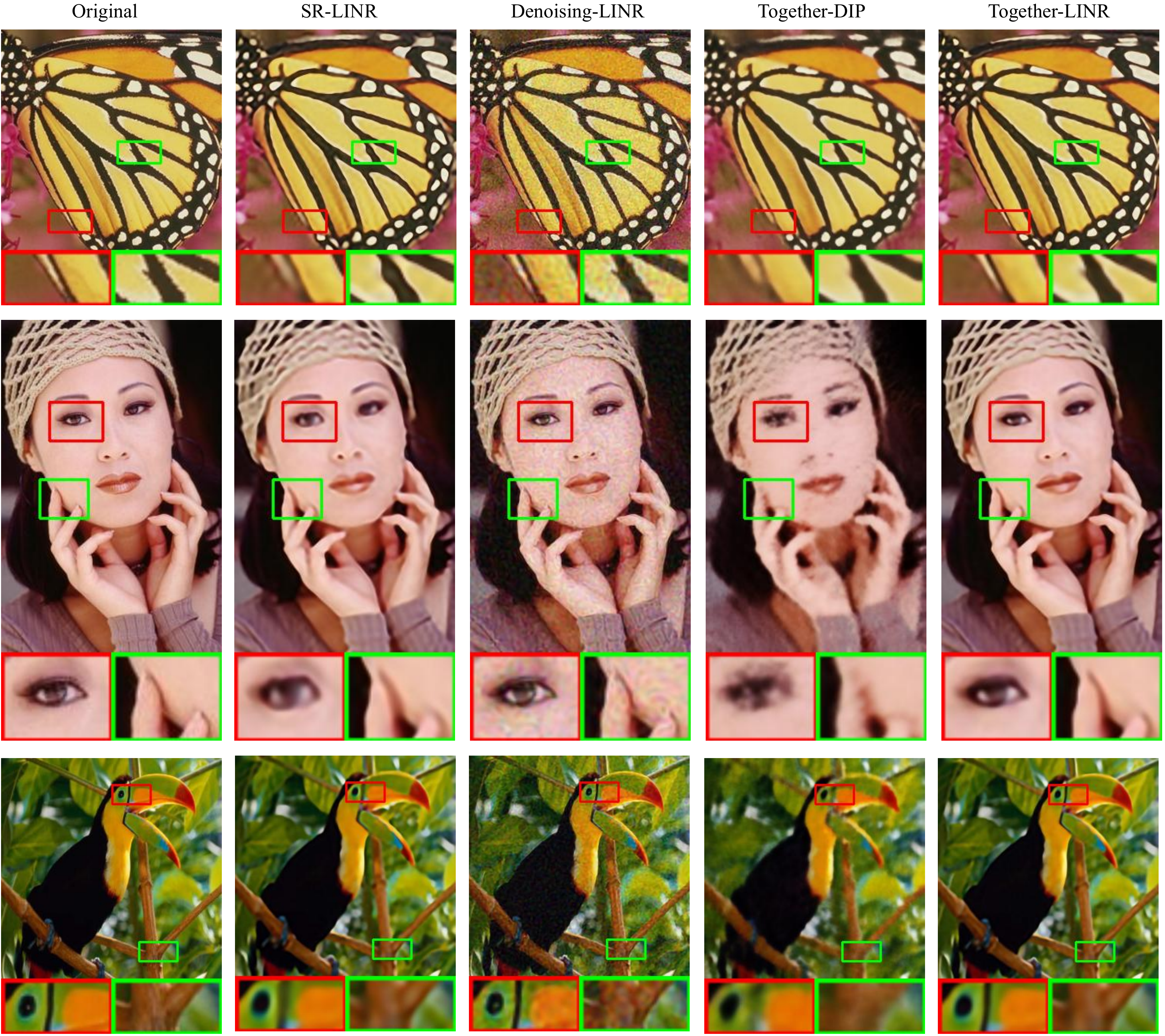}
\end{center}
\caption{The joint results on \MethodName~and DIP where both noisy and low-resolution images are used as input and the \MethodName~restoring results where only one of them is used.   }
\label{srnoise}
\end{figure}

\begin{figure}[h]
\begin{center}
\includegraphics[width=1.0\linewidth]{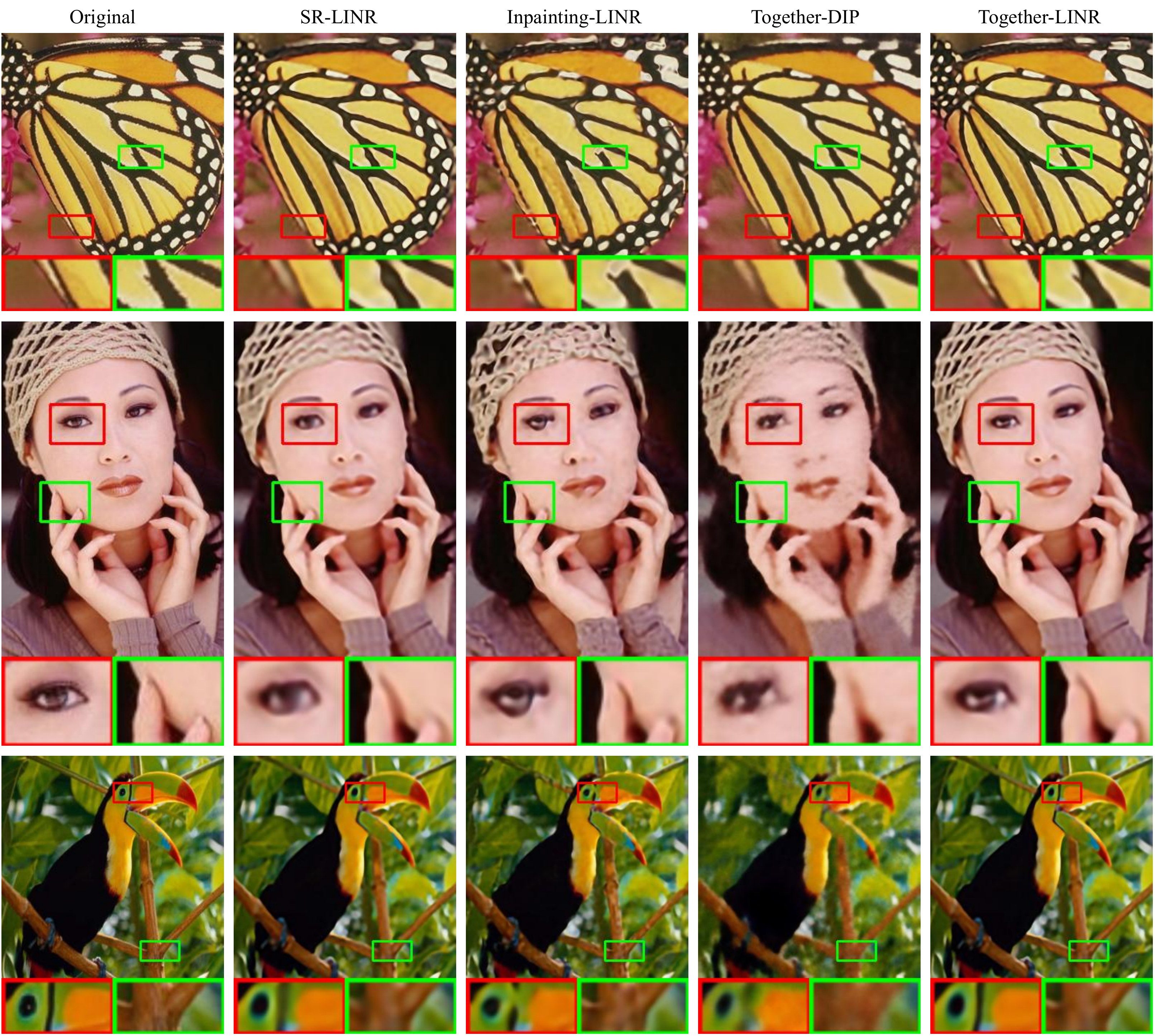}
\end{center}
\caption{The joint results on \MethodName~and DIP where both low-resolution and random masked (inpainting) images are used and the \MethodName~restoring results where only one of them is used.   }
\label{rmsr}
\end{figure}

\begin{figure}[h]
\begin{center}
\includegraphics[width=1.0\linewidth]{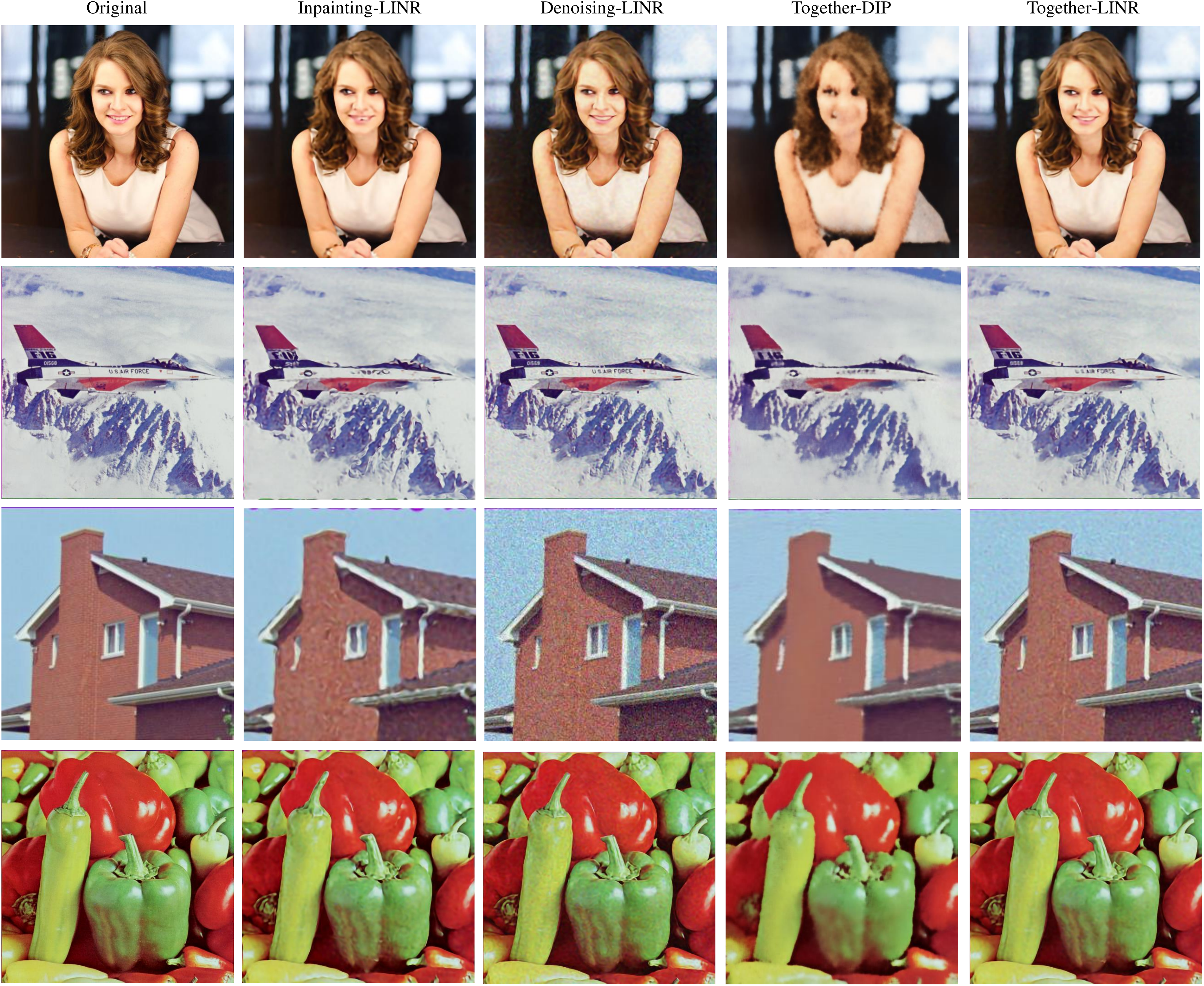}

\end{center}
\caption{The joint results on \MethodName~and DIP where both noisy and random masked (inpainting) images are used as input and the \MethodName~restoring results where only one of them is used.  }
\label{denoiserm}
\end{figure}
\FloatBarrier
\newpage
\FloatBarrier
\subsection{Qualitative results on inpainting}
\FloatBarrier
\label{sec5}
~\\~\\~\\~\\~\\~\\~\\
\begin{figure}[h]
\begin{center}
\includegraphics[width=1.0\linewidth]{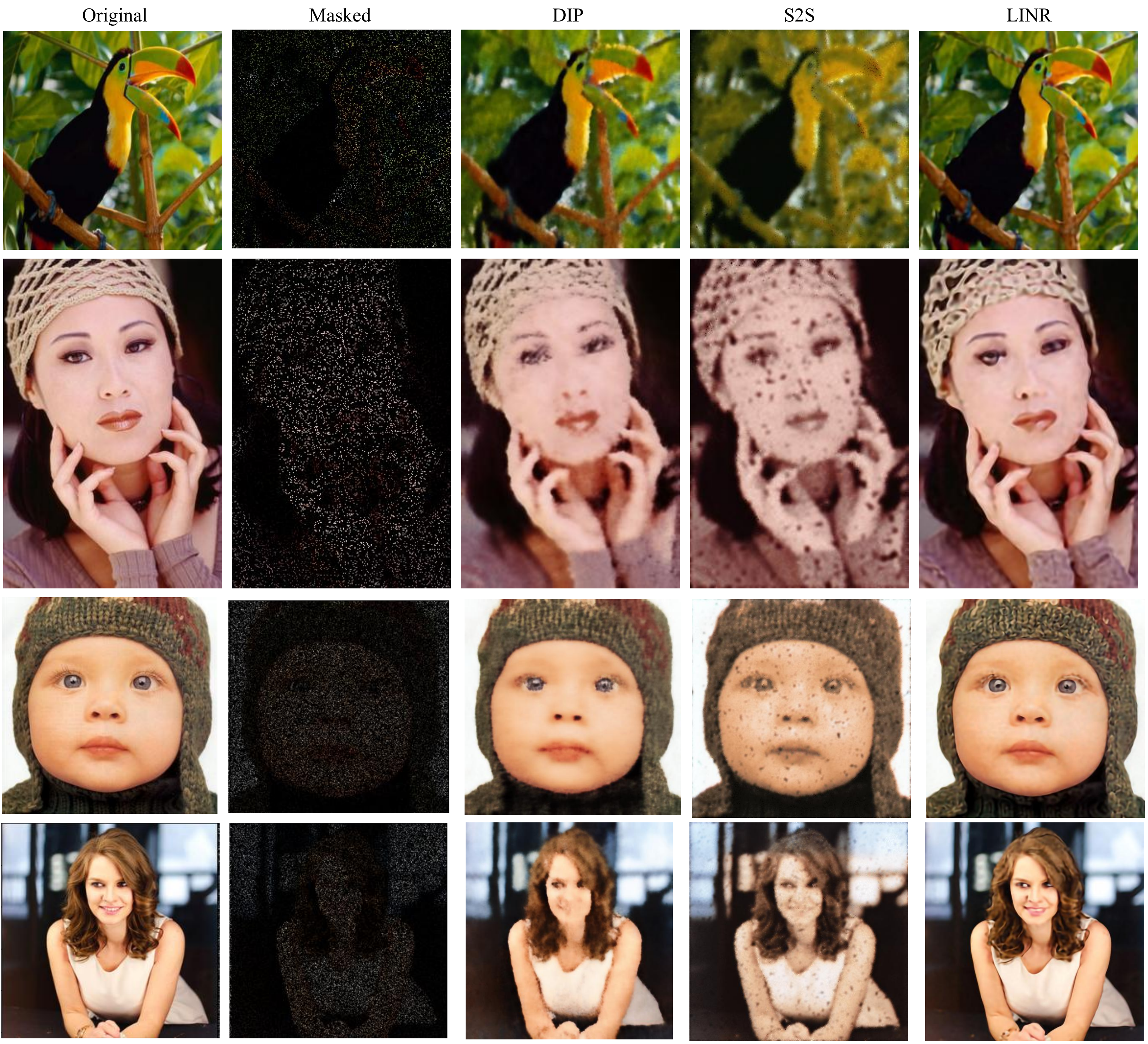}
\end{center}
\caption{Visual results on random masked inpainting with sparsity = 0.1. Even if the mask obscures most of the pixel, \MethodName~can still produce very high-quality results.}
\label{rmall}

\end{figure}
\FloatBarrier
\newpage
\subsection{Qualitative results on super-resolution}
\FloatBarrier
\label{sec6}
~\\~\\~\\~\\~\\~\\~\\
\begin{figure}[h]
\begin{center}
\includegraphics[width=1.0\linewidth]{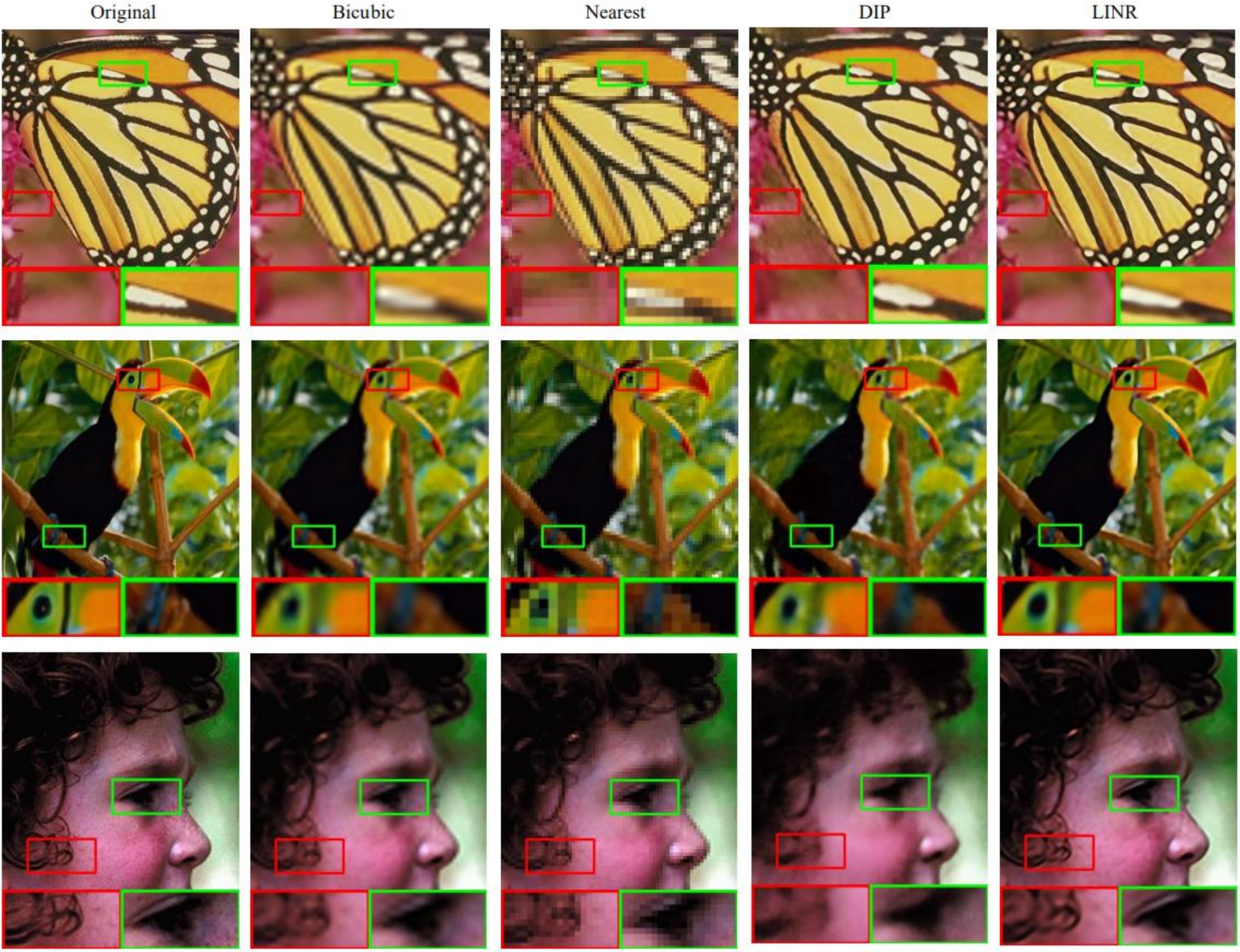}
\end{center}
\caption{Qualitative results on SR with factor = 4. Compared to other methods, \MethodName~results are clearer and contain more details with limited training resources.  }
\label{Srall}
\end{figure}
\FloatBarrier
\newpage
\subsection{Qualitative results on deblurring}
\FloatBarrier
~\\~\\~\\~\\~\\~\\~\\~\\~\\~\\
\label{sec7}
\begin{figure}[h]
\begin{center}
\includegraphics[width=0.9\linewidth]{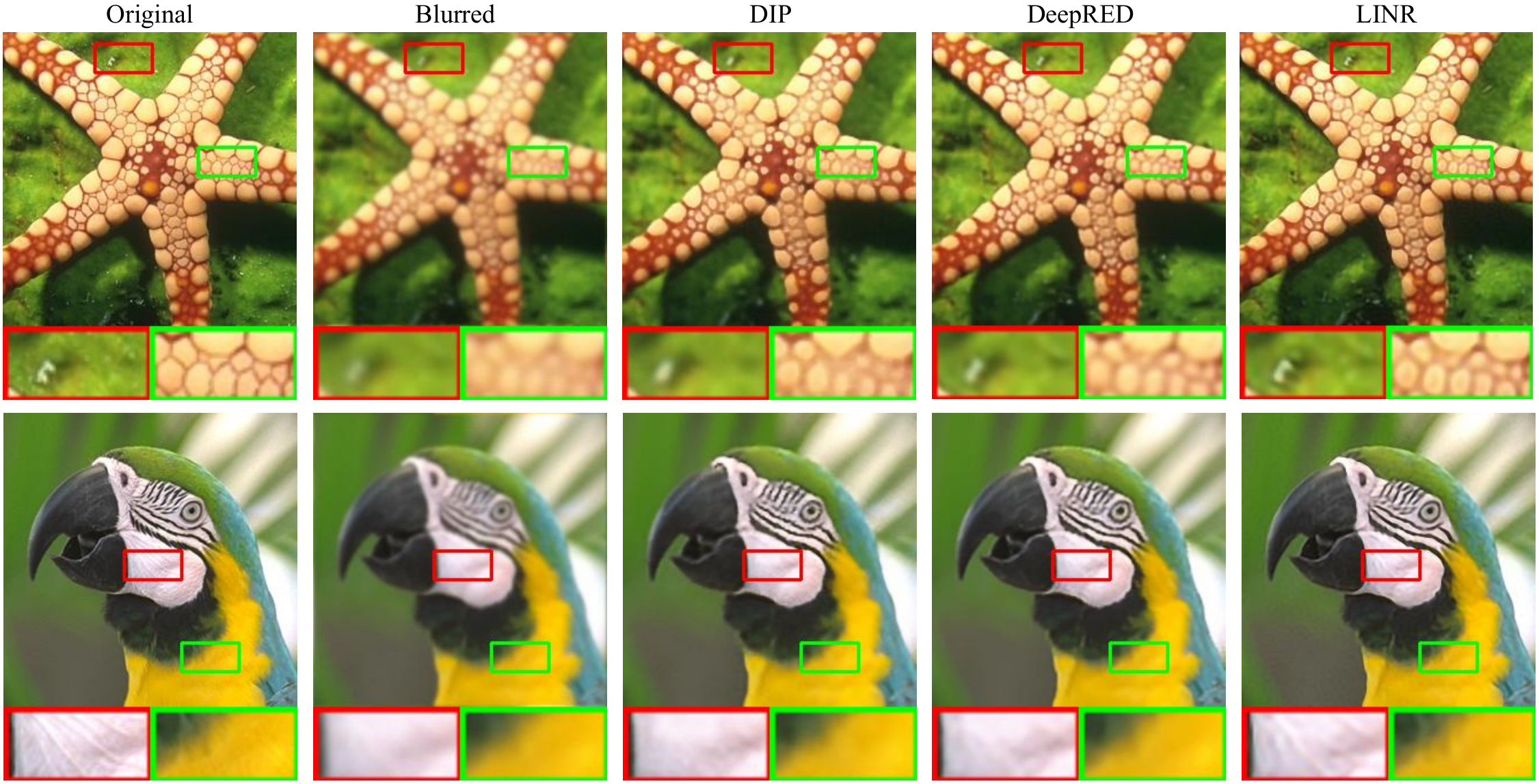}
\end{center}
\caption{Qualitative results on Gaussian blurred images. We can see that the edges of the \MethodName~are more significant and better recovered.}
\label{blurall}
\end{figure}
\FloatBarrier
\newpage
\subsection{Additional qualitative results and real noisy image denoising}
\label{sec8}
\FloatBarrier

\begin{figure}[h]
\begin{center}
\includegraphics[width=1.0\linewidth]{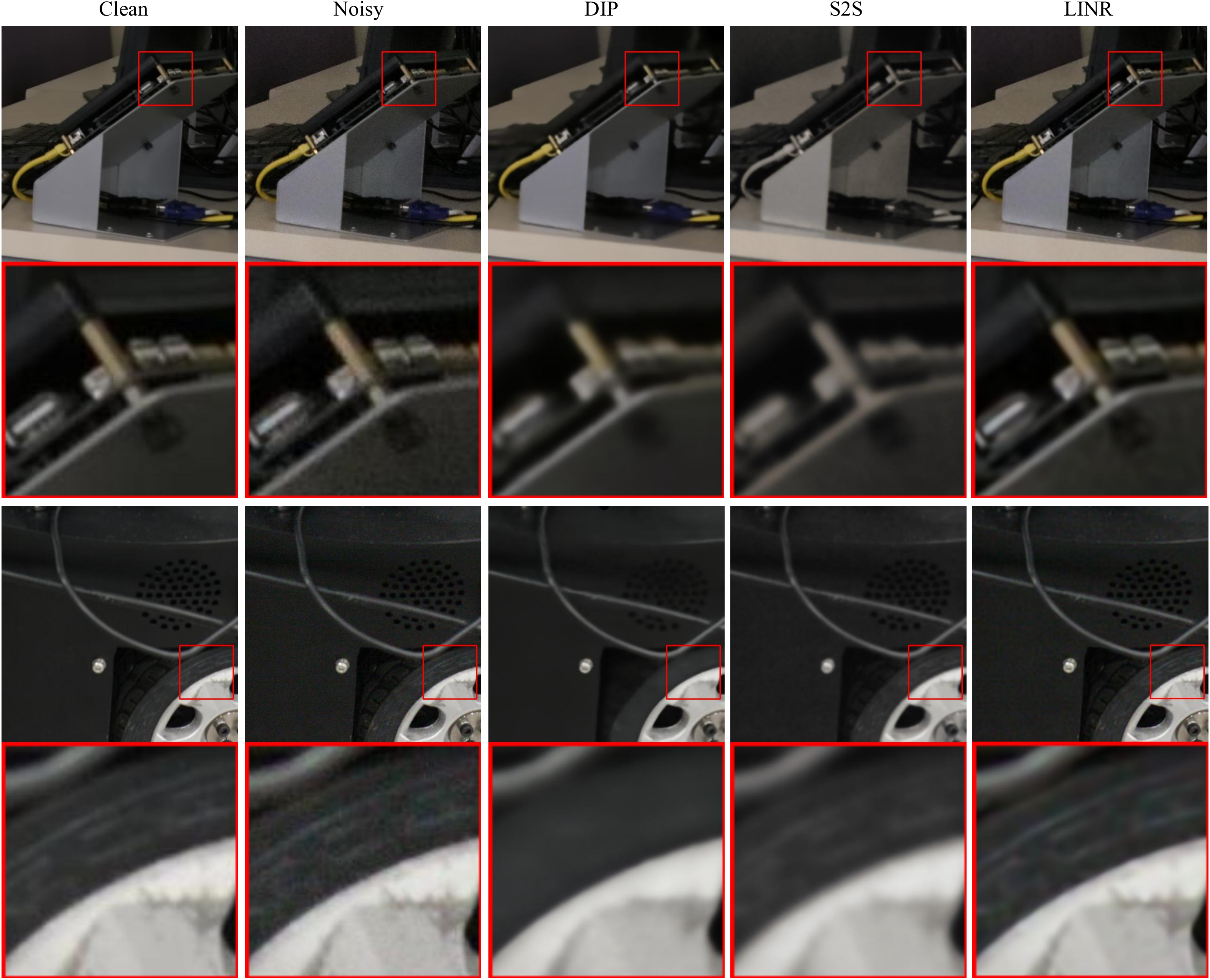}
\end{center}
\caption{Qualitative results on PolyU real-world noise dataset.}
\label{polyu}
\end{figure}

~\\~\\~\\~\\~\\
\begin{figure}[h]
\begin{center}
\includegraphics[width=1.0\linewidth]{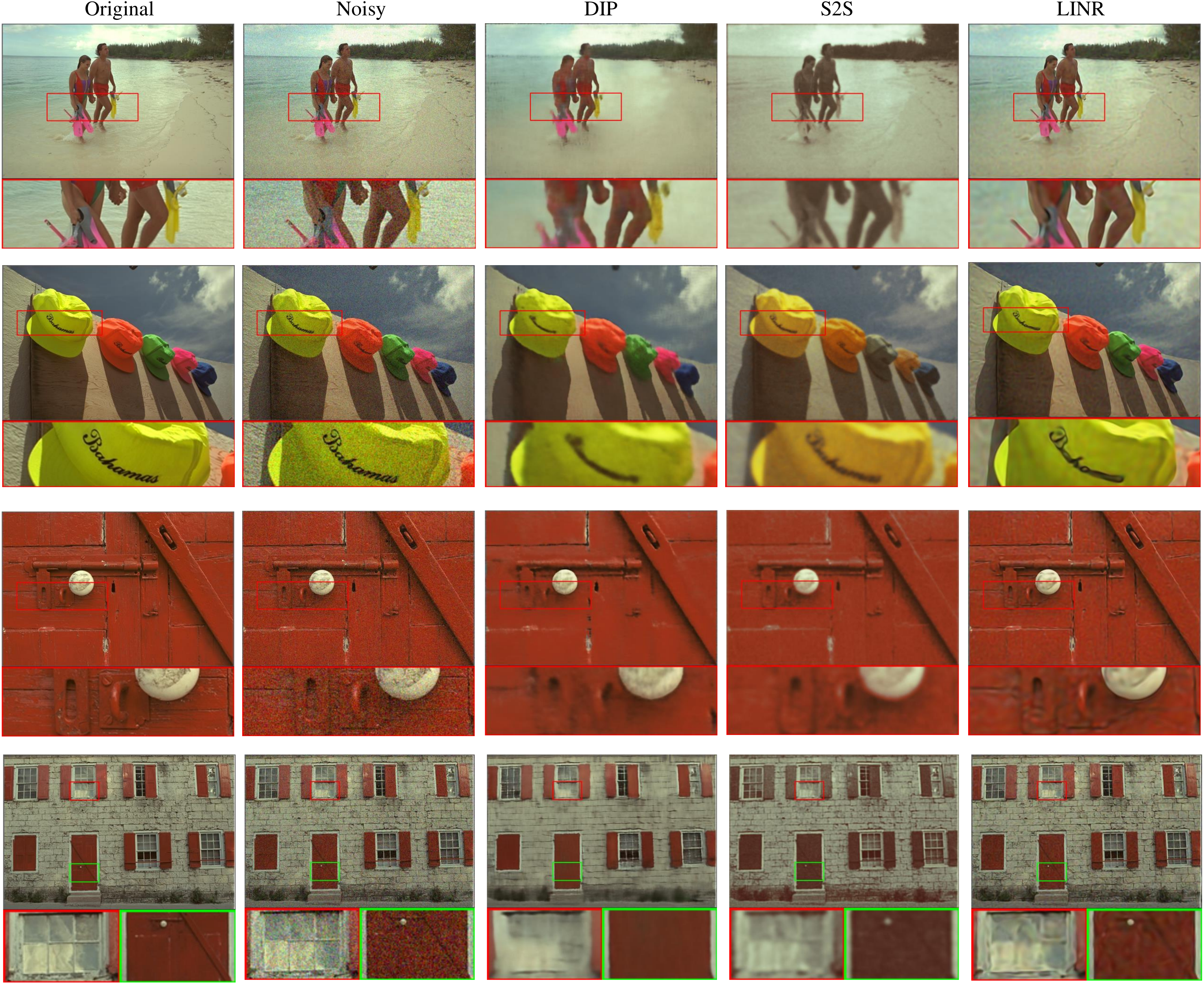}
\end{center}
\caption{Qualitative results on denoising with $\sigma = 25$.  \MethodName's denoise results contain more detailed information than other methods.}
\label{denoiseall}
\end{figure}
\FloatBarrier
\newpage
\MethodName~also works well on real-world noise. We tested the \MethodName~on PolyU \citep{xu2018real} real noisy images dataset. Due to the weak noise, we increased the training iteration for all methods to 1000. The result is shown in
\cref{denoisepolyu} and \cref{polyu}, from which we can see that the proposed \MethodName~performs better than other methods. In the experiment, we additionally used the exponential sliding window to average the result on DIP, as mentioned in their original paper \citep{ulyanov2018deep}, because, without this, DIP denoised images even perform worse (lower PSNR) than the original noisy images.

\begin{table}[htbp]
  \centering
  \caption{Average denoising performance (PSNR/SSIM) on real-world noisy images (PolyU).}
    \begin{tabular}{lccc}
    \toprule
       Images   & DIP \citep{ulyanov2018deep}   & S2S \citep{quan2020self2self}   & \MethodName \\
    \midrule
    
    PolyU & 36.07/0.968 & 32.42/0.954 & \textbf{37.12/0.979}\\
    \bottomrule
    \end{tabular}
  \label{denoisepolyu}
\end{table}

\end{document}